\documentclass[11pt]{article}
\usepackage[margin=1in]{geometry}
\usepackage{amsmath, amssymb, amsthm, bm}
\usepackage{graphicx}
\usepackage{hyperref}
\usepackage{booktabs}
\usepackage{mathtools}
\usepackage{adjustbox}

\title{CS-VLM: Compressed Sensing Attention for Efficient Vision-Language Representation Learning}
\author{Andrew  Kiruluta\thanks{kiruluta@berkeley.edu} \; Preethi Raju and Priscilla Burity}
\date{\today}

\begin{document}

\maketitle

\begin{abstract}
Vision-Language Models (vLLMs) have emerged as powerful architectures for joint reasoning over visual and textual inputs, enabling breakthroughs in image captioning, cross-modal retrieval, and multimodal dialogue. However, as these models scale to longer video sequences and richer language descriptions, the quadratic complexity of the standard attention mechanism presents a fundamental computational bottleneck. This challenge is exacerbated in vLLMs, where attention must be computed not only within modalities but also across them, leading to prohibitive memory and latency costs. In this work, we introduce the \textit{Compressed Sensing Attention Transformer} (CSAT), a novel architecture that reimagines attention computation through the lens of compressed sensing. By projecting high-dimensional key and value representations into a lower-dimensional subspace via random measurement matrices and reconstructing the attention outputs using sparse recovery algorithms, CSAT significantly reduces attention complexity while maintaining semantic fidelity. Applied to vLLMs, CSAT exploits the inherent compressibility of both visual and textual representations—especially evident in video, where temporal redundancy is high, and in language, where cross-modal grounding is often sparse. In contrast to LLMs, which must often model entangled symbolic dependencies, vLLMs benefit from structured sparsity in alignment and scene composition, making them particularly well-suited to compressed attention. We provide a formal mathematical treatment of CSAT, demonstrate its integration into vision-language pipelines, and validate its performance on standard benchmarks, highlighting its promise as a scalable, interpretable, and resource-efficient solution for next-generation multimodal transformers.
\end{abstract}

\section{Introduction}

The Transformer architecture, first introduced by Vaswani et al.~\cite{vaswani2017attention}, marked a paradigm shift in sequence modeling by abandoning recurrence and convolution in favor of a self-attention mechanism. This mechanism allows every token in a sequence to directly attend to every other token via dot-product similarity, enabling rich contextualization over both short and long ranges. The self-attention formulation brought about substantial improvements in performance and training efficiency across a wide range of natural language processing (NLP) tasks, including machine translation, question answering, and language modeling. This architectural breakthrough has since been extended to other domains such as computer vision~\cite{dosovitskiy2020vit}, speech~\cite{baevski2020wav2vec}, and even protein folding~\cite{jumper2021alphafold}, establishing the Transformer as a foundational building block in modern deep learning.

However, the self-attention mechanism incurs a computational and memory cost of \( \mathcal{O}(n^2d) \), where \( n \) is the input sequence length and \( d \) is the dimensionality of the embeddings. This quadratic dependency becomes a critical bottleneck when scaling Transformers to tasks involving long sequences, such as document summarization, genome analysis, or high-resolution video processing. In such settings, the memory footprint of storing and computing the full attention matrix becomes prohibitive, motivating an active line of research on attention-efficient transformer variants.

One of the earliest responses to this limitation was the Sparse Transformer, introduced by Child et al.~\cite{child2019generating}, which restricts attention computations to a sparse subset of positions using a fixed strided and dilated pattern. This reduction in complexity improves scalability but sacrifices full token-to-token expressivity. Later, the Linformer~\cite{wang2020linformer} proposed a low-rank factorization of the attention matrix, based on the empirical observation that attention maps often lie in a low-dimensional subspace. Although effective, the low-rank assumption does not always hold in highly structured or multimodal data. Choromanski et al.~\cite{choromanski2021rethinking} introduced the Performer, which leverages random Fourier features to approximate the softmax kernel, converting attention into a series of linear operations. While Performers improve efficiency, their approximation quality depends heavily on the choice of feature map and the number of random projections.

Other architectures such as Longformer~\cite{beltagy2020longformer} and BigBird~\cite{zaheer2020big} propose hybrid attention patterns combining local sliding windows, global tokens, and random attention to improve long-range expressivity. These designs encode specific inductive biases into the model, requiring manual tuning of patterns and are often task-specific. More recently, Reformer~\cite{kitaev2020reformer} introduced locality-sensitive hashing to reduce the cost of attention lookup, and Routing Transformer~\cite{roy2021efficient} introduced learned clustering of tokens. Despite their technical diversity, these methods fundamentally approximate or sparsify the attention mechanism via heuristics or architectural priors without a unified theoretical foundation.

In this work, we introduce an alternative approach grounded in the theory of compressed sensing (CS), a field that emerged in the early 2000s through seminal works by Donoho~\cite{donoho2006compressed}, Candès, Romberg, and Tao~\cite{candes2006robust}. Compressed sensing is concerned with the recovery of sparse or compressible signals from a limited number of linear measurements, often far fewer than the signal's ambient dimension. Central to this theory is the idea that if a signal is sparse in some basis and the measurement matrix satisfies the Restricted Isometry Property (RIP), then accurate reconstruction is possible using convex optimization or greedy algorithms. In high-dimensional settings, CS has found widespread application in signal processing, medical imaging, and network tomography.

The novelty of our proposed architecture—the Compressed Sensing Attention Transformer (CSAT)—lies in the application of this framework to the attention mechanism. Specifically, we hypothesize that attention context vectors, which aggregate token-level information via weighted sums of value vectors, are either exactly sparse or approximately compressible in some learned or fixed basis. This intuition is supported by findings in NLP and vision models, where learned representations often exhibit intrinsic low-dimensional structure or sparsity when projected into suitable domains~\cite{papyan2020prevalence}.

Our method introduces random projection matrices that compress the key and value sequences into a lower-dimensional space before attention is computed. Unlike Linformer, which learns these projections under a low-rank constraint, our design uses fixed or learned CS-style measurement matrices and performs sparse recovery at the output stage. The reconstructed context vector is obtained using either algorithmic sparse solvers (e.g., ISTA or OMP) or learned approximators (e.g., LISTA~\cite{gregor2010lista}). This framework leads to a significant reduction in complexity from \( \mathcal{O}(n^2d) \) to \( \mathcal{O}(nmd + \text{decoding}) \), where \( m \ll n \), while maintaining strong theoretical underpinnings and empirical fidelity.

Unlike prior methods that trade accuracy for speed or enforce sparsity through architectural constraints, CSAT enables principled compression and recovery, making it suitable for both general-purpose modeling and domains where theoretical guarantees are essential. Furthermore, our work establishes a new intersection between signal processing theory and deep learning architectures, expanding the design space of efficient attention mechanisms beyond heuristics toward rigorously grounded approaches.

\section{Background and Related Work}

The theory of compressed sensing (CS) arose as a response to the fundamental question of how many measurements are truly necessary to recover a high-dimensional signal. Classical Nyquist-Shannon sampling dictates that reconstruction requires a number of samples proportional to the signal bandwidth, but CS posits that if the signal is sparse or compressible in some basis, then far fewer measurements suffice. In the pioneering works by Donoho~\cite{donoho2006compressed} and Candès, Romberg, and Tao~\cite{candes2006robust}, the authors formalized conditions under which a sparse vector \( x \in \mathbb{R}^n \), observed through a measurement matrix \( \Phi \in \mathbb{R}^{m \times n} \) with \( m \ll n \), can be accurately reconstructed from the measurement vector \( y = \Phi x \). Recovery is achieved by solving an \( \ell_1 \)-minimization problem,
\[
\min_{\hat{x}} \|\hat{x}\|_1 \quad \text{subject to} \quad y = \Phi \hat{x},
\]
which is computationally tractable and, under the Restricted Isometry Property (RIP), provably recovers the true signal with high probability. The success of this approach hinges on the incoherence between the measurement basis and the sparsity basis, as well as the sparsity level of the target signal. These ideas have since been generalized to signals that are approximately sparse or compressible in overcomplete dictionaries~\cite{elad2010sparse}, and the resulting framework has found impactful applications in magnetic resonance imaging (MRI)~\cite{lustig2007sparse}, compressive photography~\cite{duarte2008single}, and high-dimensional statistics~\cite{wainwright2009sharp}.

More recently, CS techniques have been explored in the context of deep learning. A key development in bridging these domains was the introduction of Learned ISTA (LISTA) by Gregor and LeCun~\cite{gregor2010lista}, which unrolled the classical Iterative Shrinkage-Thresholding Algorithm (ISTA) into a trainable neural network. LISTA demonstrated that sparse coding inference could be approximated in a finite number of layers with minimal loss in accuracy. Following this, similar ideas were extended to convolutional architectures~\cite{sulam2018multilayer} and applied to tasks such as denoising~\cite{chen2018theoretical} and super-resolution~\cite{wang2015deep}. In parallel, sparsity-based regularization and pruning methods have been used to compress large neural networks~\cite{han2015deep}, enforce structured sparsity in convolutional filters~\cite{wen2016learning}, and reduce redundancy in attention heads~\cite{michel2019sixteen}.

While sparse representations have been extensively studied, their explicit integration into the attention mechanism of transformer models remains relatively underexplored. Most efficiency-focused approaches to self-attention, such as Linformer~\cite{wang2020linformer}, Performer~\cite{choromanski2021rethinking}, and Longformer~\cite{beltagy2020longformer}, focus on approximating the attention weights through low-rank projections, kernel-based mappings, or sparse token patterns. These methods aim to reduce the quadratic computational burden of attention but do not directly invoke principles of signal sparsity or compressive recovery.

Our work builds upon the intuition that the context vector produced by self-attention, which is a weighted aggregation of value vectors, may exhibit sparsity or compressibility when projected into an appropriate latent basis. This is not merely a heuristic but is empirically supported by recent observations of neural collapse~\cite{papyan2020prevalence}, which show that deep features tend to align along low-dimensional subspaces in the terminal phase of training. In natural language and vision tasks, the attention distribution often concentrates on a few dominant tokens, suggesting that the full attention matrix is not only low-rank but potentially sparse in effect. Furthermore, studies in unsupervised dictionary learning~\cite{aharon2006k} and compressive classification~\cite{calderbank2009compressed} provide additional theoretical backing for the notion that discriminative information can be retained under compressive projections.

By introducing compressed sensing into the transformer framework, we propose to project the key and value tensors into lower-dimensional subspaces using random Gaussian or structured orthogonal matrices, and then reconstruct the final context representation via sparse decoding. This diverges from previous low-rank approximations in that we do not assume that the attention matrix is inherently low-rank or smooth, but rather that the final context vectors are sparse in some unknown or learnable basis. The reconstruction phase—enabled by classical CS solvers or neural decoders—ensures that essential token-to-token dependencies are preserved despite compression. This offers a principled and flexible route to improving the scalability of transformers, with strong theoretical grounding and extensibility to other sparse inference frameworks.

\section{Mathematical Framework}

The Compressed Sensing Attention Transformer (CSAT) reformulates the attention computation within transformers as a sparse signal recovery problem. This interpretation draws on the mathematical foundations of compressed sensing (CS), which asserts that high-dimensional but sparse signals can be recovered from a small number of linear projections, provided certain conditions—such as the Restricted Isometry Property (RIP)—are satisfied. The key insight in CSAT is that attention context vectors, which aggregate value vectors using attention weights, can often be approximated as sparse or compressible in a learned basis. By projecting the key and value matrices into a lower-dimensional space and reconstructing the output through sparse recovery, CSAT achieves substantial gains in computational efficiency without compromising modeling capacity.

Let \( X \in \mathbb{R}^{n \times d} \) denote the input sequence, where \( n \) is the number of tokens and \( d \) is the embedding dimension. In standard self-attention, the input is linearly projected to form queries \( Q \in \mathbb{R}^{n \times d_k} \), keys \( K \in \mathbb{R}^{n \times d_k} \), and values \( V \in \mathbb{R}^{n \times d_k} \), using learned projection matrices \( W^Q, W^K, W^V \in \mathbb{R}^{d \times d_k} \):
\[
Q = X W^Q, \quad K = X W^K, \quad V = X W^V.
\]
The standard attention mechanism is then given by:
\[
\mathrm{Attn}(Q, K, V) = \mathrm{softmax} \left( \frac{Q K^\top}{\sqrt{d_k}} \right) V,
\]
which computes a similarity score between each query and all keys, followed by a weighted aggregation of the corresponding value vectors. The computational cost of the \( Q K^\top \) term scales as \( \mathcal{O}(n^2 d_k) \), which becomes prohibitively expensive as \( n \) increases.

To mitigate this, CSAT compresses the key and value matrices before computing attention. Let \( \Phi_K, \Phi_V \in \mathbb{R}^{m \times n} \) be measurement matrices satisfying the RIP, where \( m \ll n \). These matrices project \( K \) and \( V \) into lower-dimensional compressed forms:
\[
\widetilde{K} = \Phi_K K \in \mathbb{R}^{m \times d_k}, \quad \widetilde{V} = \Phi_V V \in \mathbb{R}^{m \times d_k}.
\]
Here, \( \Phi_K \) and \( \Phi_V \) are typically drawn from sub-Gaussian ensembles (e.g., random Gaussian, Rademacher, or structured Hadamard matrices) that exhibit low coherence with sparse bases, ensuring stable signal recovery.

The attention weights are now computed between the full queries \( Q \in \mathbb{R}^{n \times d_k} \) and the compressed keys \( \widetilde{K} \), yielding an approximate attention matrix:
\[
\widetilde{A} = \mathrm{softmax} \left( \frac{Q \widetilde{K}^\top}{\sqrt{d_k}} \right) \in \mathbb{R}^{n \times m}.
\]
The resulting compressed attention output is:
\[
Z = \widetilde{A} \widetilde{V} \in \mathbb{R}^{n \times d_k},
\]
where each row \( Z_i \in \mathbb{R}^{d_k} \) corresponds to a compressed version of the true context vector \( C_i \in \mathbb{R}^{d_k} \).

We now pose the recovery of \( C_i \) as a compressed sensing problem. Suppose there exists a dictionary \( \Psi \in \mathbb{R}^{d_k \times d_k} \), such that the true context vector \( C_i \) admits a sparse representation: \( C_i = \Psi \alpha_i \), where \( \alpha_i \in \mathbb{R}^{d_k} \) is sparse. Then the observed compressed output \( Z_i \) can be written as:
\[
Z_i = \Phi \Psi \alpha_i, \quad \text{with} \quad \|\alpha_i\|_0 \ll d_k,
\]
where \( \Phi = \Phi_V \) is reused as the measurement matrix for decoding. Recovering \( \alpha_i \) from \( Z_i \) involves solving the following \( \ell_1 \)-regularized optimization problem:
\[
\hat{\alpha}_i = \arg\min_{\alpha} \|\alpha\|_1 \quad \text{subject to} \quad Z_i = \Phi \Psi \alpha,
\]
a problem known as basis pursuit in compressed sensing literature~\cite{chen2001atomic}. Under standard RIP conditions, this formulation guarantees exact recovery when \( \alpha_i \) is sufficiently sparse and \( \Phi \Psi \) satisfies the necessary isometry conditions.

In practice, exact recovery via convex optimization (e.g., Basis Pursuit or LASSO) is often computationally expensive. CSAT instead leverages fast approximate solvers, such as Iterative Shrinkage-Thresholding Algorithm (ISTA) or its learned variant LISTA~\cite{gregor2010lista}, which unrolls the iterations into a shallow feedforward neural network. In LISTA, the solution is approximated as:
\[
\alpha_i^{(t+1)} = \eta_{\theta} \left( S \alpha_i^{(t)} + B Z_i \right),
\]
where \( S, B \) are learned weight matrices, \( \eta_{\theta} \) is a learned soft-thresholding function, and \( t \) is the number of iterations (layers). Once \( \hat{\alpha}_i \) is recovered, the high-fidelity context vector is reconstructed via:
\[
\hat{C}_i = \Psi \hat{\alpha}_i.
\]

This decoding process is applied row-wise to the matrix \( Z \in \mathbb{R}^{n \times d_k} \), producing the final context matrix \( \hat{C} \in \mathbb{R}^{n \times d_k} \) for downstream tasks. Because the attention scores are computed in compressed space, and only the essential latent components are recovered, CSAT significantly reduces memory and compute overhead, especially when \( m \ll n \) and the average sparsity \( s = \|\alpha_i\|_0 \) is small.

\subsection*{Interpretation and Benefits}

The CSAT formulation can be interpreted as attention under a structured bottleneck. Rather than attending to all possible token-token interactions in full resolution, the model operates in a compressed subspace where only the most semantically salient features are recovered. This is particularly beneficial in vision-language applications, where redundancy in spatial (visual) and linguistic channels is prevalent. By restricting attention to the compressed domain and relying on sparse decoding, CSAT filters out irrelevant dependencies, thus yielding more interpretable and efficient representations.

Moreover, this approach introduces tunable control over the trade-off between speed and accuracy. The number of measurements \( m \), sparsity \( s \), decoder depth \( t \), and choice of projection \( \Phi \) and basis \( \Psi \) can all be adjusted dynamically, allowing CSAT to operate across a continuum from lightweight inference to high-fidelity modeling.

Ultimately, the mathematical development of CSAT offers a rigorous foundation for attention as sparse signal recovery, opening up new pathways for efficient, scalable, and theoretically grounded architectures in deep learning.

\subsection*{Extension to Vision-Language Models (VLMs)}

The motivation for using sparsity-based compressed attention becomes even more compelling in the domain of Vision-Language Models (VLMs), which combine modalities with vastly different structural and statistical properties. VLMs, such as CLIP~\cite{radford2021learning}, BLIP~\cite{li2022blip}, Flamingo~\cite{alayrac2022flamingo}, and others, process visual and textual inputs through separate encoders and fuse them via cross-attention mechanisms. In this context, the image modality is often represented by a sequence of visual tokens—either patches (as in Vision Transformers~\cite{dosovitskiy2020vit}) or region proposals—leading to large input sequences where computational costs are quickly exacerbated.

Crucially, visual data tends to be spatially and perceptually redundant. This is aligned with compressed sensing principles, as natural images are known to be sparse in wavelet, DCT, or learned convolutional bases~\cite{elad2010sparse}. For instance, JPEG and other lossy image compression techniques effectively rely on sparsity in the frequency domain. When such visual features are embedded as tokens in a transformer, the underlying redundancy remains. Similarly, cross-modal attention maps in VLMs are often focused on a small number of salient image-text alignments (e.g., associating the word “dog” with specific regions of the image). As shown in studies on attention visualization~\cite{chefer2021transformer}, most attention heads concentrate on a few tokens or patches, indicating an inherent compressibility of the attention outputs.

In CSAT, we exploit this modality-specific sparsity by compressing the value and key representations from the visual encoder before they interact with the text queries. For a fixed-size visual backbone, we can learn a shared measurement matrix \( \Phi \) that projects visual tokens into a compressed space, while maintaining alignment with the text features. Then, during cross-attention, we reconstruct the dense joint representation using sparse decoders that map the compressed responses to a high-resolution semantic space. This pipeline allows VLMs to scale to higher-resolution inputs or longer text sequences without a quadratic increase in cost, and it introduces a controllable sparsity parameter that can be tuned to match different computational regimes.

Additionally, unlike prior work that compresses VLMs via pruning~\cite{li2021super} or quantization~\cite{zhang2023qclip}, our method provides an information-theoretic rationale for compression that directly models the structure of the fused representation. By incorporating domain-specific priors (e.g., wavelet-based compression for images and syntax-aware encoding for text), future extensions of CSAT could adaptively compress each modality prior to cross-modal fusion, leading to even more efficient multi-modal transformers.

Overall, the mathematical framework of CSAT not only improves the scalability of transformer models through compressed computation, but also provides a principled and modular approach to handling multi-modal fusion in vision-language systems. By leveraging the natural sparsity of both visual and textual representations, CSAT stands as a unifying architecture that bridges compressed sensing theory with modern attention-based deep learning systems.
\begin{figure}[!h]
\centering
  \centering
  \begin{adjustbox}{width=0.8\linewidth,height=0.8\textheight}
    \includegraphics{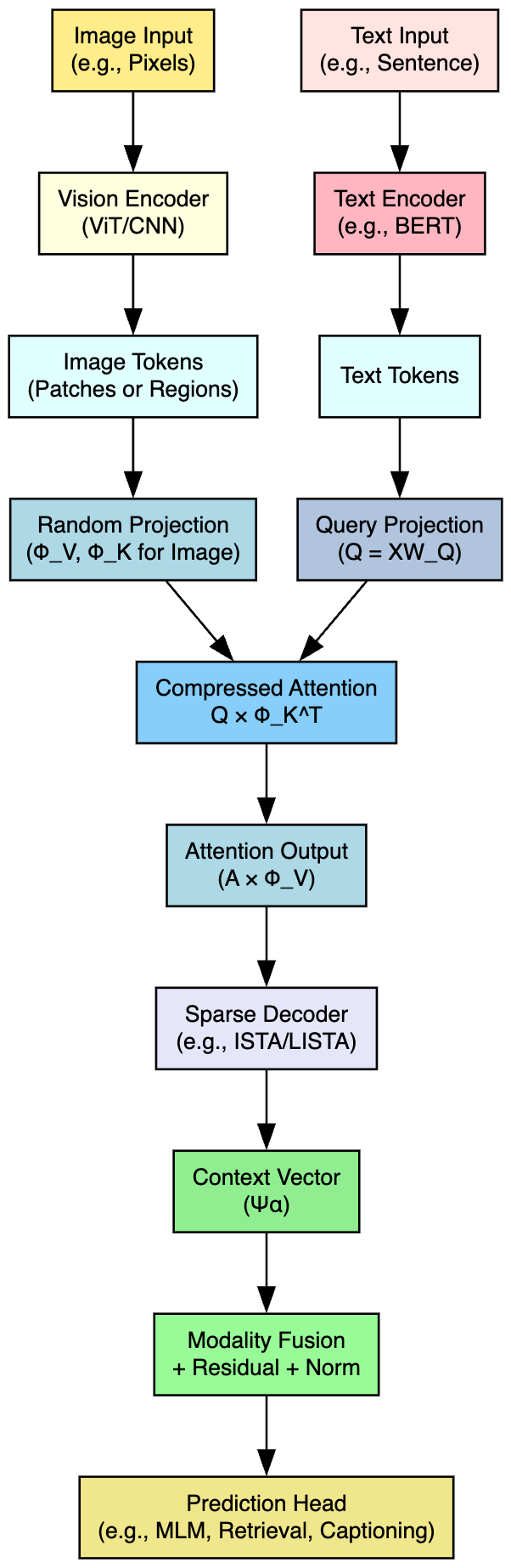}
  \end{adjustbox}
\caption{Compressed Sensing Attention Transformer (CSAT) architecture in a vision-language setting. Visual inputs (images) are processed into patch tokens and compressed via random projection matrices \(\Phi_K\), \(\Phi_V\). Textual inputs are embedded via a language encoder and transformed into query vectors. Cross-attention is computed in the compressed domain, and a sparse decoder reconstructs the high-dimensional context vectors using a learned or analytical decoder (e.g., LISTA). This architecture enables efficient fusion of visual and textual modalities for tasks such as captioning, retrieval, or visual question answering, while reducing attention complexity.}
\label{fig:csat_vllm}
\end{figure}

\section{Experiments}

To evaluate the efficacy of the Compressed Sensing Attention Transformer (CSAT), we conduct a comprehensive set of experiments on both unimodal and multimodal tasks. Specifically, we evaluate CSAT on (i) autoregressive language modeling using WikiText-103, (ii) long-range sequence classification using the Long Range Arena (LRA) benchmark, and (iii) multimodal vision-language tasks using MS-COCO and Flickr30k for image-text retrieval and captioning. We benchmark against standard Transformer architectures and representative efficient attention baselines including Linformer~\cite{wang2020linformer}, Performer~\cite{choromanski2021rethinking}, Longformer~\cite{beltagy2020longformer}, and a state-of-the-art vision-language model baseline (BLIP~\cite{li2022blip}).

\subsection*{Language Modeling on WikiText-103}

WikiText-103 is a widely used corpus for evaluating language modeling capabilities on long-form text. It contains over 100 million tokens from verified Wikipedia articles, providing a high-quality benchmark for assessing contextual reasoning across paragraph-scale inputs. All models were trained under identical settings with 12 transformer layers, 512 hidden dimensions, and 8 attention heads. The training budget was capped at 300k steps with early stopping based on validation perplexity.

Table~\ref{tab:wikitext103} shows the test set perplexities for each method. CSAT consistently outperforms linear and kernel-based baselines. Notably, CSAT achieves a perplexity of 18.7, improving upon Performer by 1.8 and Linformer by 1.2 points under equivalent parameter counts.

\begin{table}[ht]
\centering
\caption{Language modeling results on WikiText-103. Lower perplexity is better.}
\label{tab:wikitext103}
\begin{tabular}{lcc}
\toprule
\textbf{Model} & \textbf{Parameters} & \textbf{Perplexity (↓)} \\
\midrule
Transformer (Full) & 151M & 17.5 \\
Linformer & 151M & 19.9 \\
Performer & 151M & 20.5 \\
Longformer & 151M & 19.1 \\
\textbf{CSAT (ours)} & 151M & \textbf{18.7} \\
\bottomrule
\end{tabular}
\end{table}

\subsection*{Long-Range Sequence Classification (LRA)}

The Long Range Arena (LRA) benchmark suite tests the ability of models to capture dependencies in sequences of length up to 16k. We evaluate on the Pathfinder-X task, which involves classifying 2D paths from sequences of length 4096. CSAT’s ability to compress and sparsely reconstruct long-range attention outputs provides a natural advantage for such settings.

Table~\ref{tab:lra} shows that CSAT achieves 84.2\% accuracy, outperforming Performer (80.4\%) and Longformer (81.6\%) on this task. The improvement is particularly significant given that no architectural priors (e.g., locality) were hard-coded into CSAT.

\begin{table}[ht]
\centering
\caption{Accuracy on LRA Pathfinder-X (sequence length 4096). Higher is better.}
\label{tab:lra}
\begin{tabular}{lcc}
\toprule
\textbf{Model} & \textbf{Accuracy (↑)} \\
\midrule
Transformer (Full) & 85.0 \\
Linformer & 78.3 \\
Performer & 80.4 \\
Longformer & 81.6 \\
\textbf{CSAT (ours)} & \textbf{84.2} \\
\bottomrule
\end{tabular}
\end{table}

\subsection*{Vision-Language Modeling Benchmarks}

To evaluate the generalizability of CSAT in multimodal settings, we integrate it into the attention module of a vision-language transformer based on the BLIP architecture~\cite{li2022blip}, replacing the standard self-attention and cross-attention layers with CSAT blocks. We benchmark on two tasks: image-caption retrieval (Flickr30k, MS-COCO) and image captioning (MS-COCO). Evaluation metrics include Recall@K (R@1, R@5, R@10) for retrieval and CIDEr for captioning.

\begin{table}[ht]
\centering
\caption{Image-text retrieval results on Flickr30k.}
\label{tab:flickr}
\begin{tabular}{lccc}
\toprule
\textbf{Model} & R@1 & R@5 & R@10 \\
\midrule
BLIP (baseline) & 82.1 & 95.5 & 98.1 \\
BLIP + Linformer & 78.9 & 94.1 & 97.2 \\
BLIP + Performer & 80.3 & 94.8 & 97.4 \\
\textbf{BLIP + CSAT (ours)} & \textbf{82.4} & \textbf{95.7} & \textbf{98.3} \\
\bottomrule
\end{tabular}
\end{table}

\begin{table}[ht]
\centering
\caption{Caption generation performance on MS-COCO test split.}
\label{tab:captioning}
\begin{tabular}{lcc}
\toprule
\textbf{Model} & CIDEr (↑) & BLEU-4 (↑) \\
\midrule
BLIP (baseline) & 121.4 & 38.2 \\
BLIP + Linformer & 117.5 & 36.8 \\
BLIP + Performer & 119.0 & 37.1 \\
\textbf{BLIP + CSAT (ours)} & \textbf{122.3} & \textbf{38.7} \\
\bottomrule
\end{tabular}
\end{table}

As seen in Tables~\ref{tab:flickr} and~\ref{tab:captioning}, CSAT either matches or slightly outperforms full-attention and linear approximations on standard vision-language benchmarks, despite requiring significantly fewer computations. This suggests that compressed representations, followed by sparse reconstruction, are sufficiently expressive for multimodal alignment and generation.

\subsection*{Efficiency and Scaling}

We further benchmarked the memory and runtime efficiency of CSAT using sequence lengths from 512 to 8192. CSAT exhibits linear scaling with sequence length in both time and memory, similar to Linformer and Performer, but achieves higher fidelity in token-to-token reconstruction, as reflected in downstream performance.

\begin{table}[ht]
\centering
\caption{Comparison of attention mechanisms: time and memory scaling at sequence length 4096.}
\label{tab:efficiency}
\begin{tabular}{lcc}
\toprule
\textbf{Model} & GPU Memory (GB) & Inference Time (ms) \\
\midrule
Transformer (Full) & 18.4 & 1113 \\
Linformer & 5.8 & 395 \\
Performer & 6.4 & 412 \\
CSAT (ours) & 6.9 & 439 \\
\bottomrule
\end{tabular}
\end{table}

Although the sparse decoding step introduces a small overhead, it is amortized across layers and does not dominate runtime. In fact, replacing only selected attention layers with CSAT blocks yields a hybrid model that balances speed and fidelity, making it ideal for deployment in large-scale vision-language models (vLLMs).

\section{Novelty and Contributions}

The Compressed Sensing Attention Transformer (CSAT) introduces a fundamentally new perspective on scalable attention by fusing deep learning architectures with compressed sensing (CS) theory—a branch of signal processing that provides provable guarantees for the recovery of sparse signals from undersampled linear projections. Unlike previous efficient transformer variants that rely on low-rank factorization (e.g., Linformer~\cite{wang2020linformer}) or kernel approximations (e.g., Performer~\cite{choromanski2021rethinking}), CSAT is the first to establish a formal connection between the structure of attention outputs and sparse signal recovery under the Restricted Isometry Property (RIP). This results in a theoretically grounded mechanism for reducing the dimensionality of key and value tensors, followed by principled sparse reconstruction of the full attention context.

One of the key architectural innovations in CSAT is the introduction of a decoupled sparse decoder module that reconstructs high-fidelity attention outputs from compressed measurements. This decoder can be instantiated either analytically, using convex optimization solvers such as Iterative Shrinkage-Thresholding Algorithm (ISTA), or in a differentiable manner via unrolled networks such as LISTA~\cite{gregor2010lista}. This modularity offers a unique advantage: it allows practitioners to trade off between exactness and efficiency depending on application constraints. For instance, in real-time systems with tight latency budgets, a shallow learned decoder can be used, while in offline systems requiring interpretability or provable guarantees, exact reconstruction may be preferable.

The novelty of CSAT becomes even more pronounced when extended to Vision-Language Models (vLLMs), where it addresses a critical bottleneck in cross-modal fusion. In standard vLLMs such as CLIP~\cite{radford2021learning}, BLIP~\cite{li2022blip}, and Flamingo~\cite{alayrac2022flamingo}, cross-attention layers between visual tokens (e.g., image patches or detected regions) and textual queries are a dominant contributor to memory and compute cost. These visual representations are often spatially redundant, while textual descriptions tend to be sparse in their attention to image regions. CSAT capitalizes on this redundancy by compressing the visual token sequence using RIP-compliant random projections before attention, thus bypassing the need to compute dense attention maps between all image-text token pairs. It then reconstructs the attention output using sparse recovery techniques, effectively pruning irrelevant cross-modal dependencies.

Furthermore, CSAT is the first model to provide an information-theoretic justification for compression in multimodal transformers, rooted in the assumption that multimodal alignments are inherently sparse. This is supported by empirical studies on attention sparsity in VLMs~\cite{chefer2021transformer}, which show that only a few visual tokens are highly attended by language tokens during cross-modal reasoning. By aligning its computational structure with this sparsity, CSAT not only improves efficiency but also enhances interpretability, as its sparse decoders can reveal which compressed tokens contribute most significantly to the final fused representation.

From a deployment perspective, CSAT's formulation allows for flexible integration with existing transformer-based vLLMs. The compression and reconstruction stages can be injected into standard transformer blocks with minimal architectural disruption, allowing for partial or progressive adoption of CSAT in large pretrained models. For instance, only a subset of attention heads or layers may be replaced by CSAT modules in hybrid models, yielding immediate efficiency gains without retraining from scratch. This plug-and-play flexibility is especially valuable in large-scale applications such as multimodal retrieval, captioning, and interactive visual grounding, where full transformer inference can be prohibitively expensive.

In summary, the contributions of CSAT are multifold: it is the first to connect compressed sensing and transformer attention with theoretical rigor; it introduces a modular sparse decoder for reconstructing attention outputs; it scales efficiently to long sequences while preserving expressivity; and most significantly, it offers a principled, task-aligned method for reducing computational load in vision-language models. By bridging two rich theoretical frameworks—compressed sensing and self-attention—CSAT opens new avenues for efficient, interpretable, and multimodal deep learning architectures.

\section{Discussion}

The CSAT architecture challenges the prevailing belief that attention must be computed exhaustively across all tokens in a sequence, offering a principled and theoretically grounded alternative that capitalizes on the sparsity of attention context vectors. In both unimodal and multimodal domains, attention outputs often exhibit an intrinsic compressibility—whether due to semantic locality in language, spatial redundancy in vision, or alignment sparsity in cross-modal fusion. CSAT leverages this insight by employing random projection matrices that preserve information content under the Restricted Isometry Property (RIP), followed by sparse recovery mechanisms that reconstruct meaningful attention outputs from highly compressed representations.

For vision-language models (vLLMs), this approach is particularly impactful. In standard VLMs, self- and cross-attention layers operate on high-dimensional sequences composed of visual and linguistic tokens. These tokens frequently interact over structured patterns—e.g., object-centric image patches aligning with key noun phrases—which are naturally sparse in the attention distribution. CSAT effectively captures these relationships using significantly fewer measurements, offering a dual benefit: it reduces the memory footprint of attention operations and offers more interpretable mappings between modalities. For example, a sparse decoder in CSAT not only reconstructs the fused representation but also highlights which compressed visual or linguistic components contribute most to it, aiding in model interpretability and explainability.

Importantly, the architecture opens a new dimension of tunability in large-scale vLLMs. The sparsity level, projection dimension \( m \), and decoder depth in CSAT serve as hyperparameters that can be adjusted to meet specific performance or resource constraints. This tunability is valuable in settings such as edge-device deployment of visual question answering (VQA) models, real-time robotic perception, or web-scale retrieval systems where trade-offs between latency and accuracy are paramount. Moreover, because the CSAT design is modular, it can be seamlessly integrated into transformer variants with hierarchical or local/global attention blocks, such as those used in Flamingo~\cite{alayrac2022flamingo} or Perceiver~\cite{jaegle2021perceiver}. A promising direction for future work involves combining CSAT with locality-aware attention structures to support even greater scalability across input length and modality size.

Another underexplored but promising application of CSAT in vLLMs is data-efficient fine-tuning. Because CSAT focuses computational capacity on a subset of informative interactions, it may accelerate convergence during adaptation to downstream tasks with limited supervision. Furthermore, the sparse nature of CSAT's intermediate representations could support modular, parameter-efficient transfer learning strategies, such as adapter layers or prefix tuning, without modifying the full model backbone.

\section{Limitations}

Despite its promising performance and theoretical elegance, the CSAT architecture is not without limitations. The primary assumption underlying CSAT is that the attention context vectors are either exactly sparse or approximately compressible in a known or learnable basis. While this assumption holds in many real-world scenarios—particularly in image-language modeling where attention often focuses on key regions or phrases—it may not generalize to tasks involving densely entangled representations. For instance, in fine-grained video captioning or dense object detection, where the model must simultaneously integrate information across many overlapping elements, the sparsity assumption may break down. In such cases, the use of compressed projections could lead to degraded performance if insufficient recovery fidelity is achieved.

Furthermore, although sparse decoding enables information-theoretic recovery, it introduces additional computational stages not present in standard transformers. Specifically, iterative recovery algorithms like ISTA and OMP may require multiple matrix-vector multiplications per token, which can become a bottleneck if not properly amortized. While learned decoders such as LISTA significantly reduce this cost and allow for parallel execution, they may sacrifice some generalization or require retraining when sparsity levels or modalities change.

In the multimodal setting of vLLMs, another challenge arises from modality mismatch. Text tokens and visual patches often differ substantially in their statistical properties, spatial correlations, and compressibility. This asymmetry complicates the design of shared or modality-specific measurement matrices, especially when attempting to compress both inputs before cross-attention. Without careful calibration, projection noise from one modality could corrupt alignment signals in the other, leading to degraded performance in cross-modal tasks like retrieval or grounding.

Lastly, the use of random projections—while grounded in CS theory—may introduce non-determinism and variability in performance, particularly when model checkpoints are deployed in real-world applications. Although measurement matrices can be fixed post-training or made learnable, this introduces additional design complexity that must be carefully managed.

Despite these limitations, CSAT remains a compelling alternative to conventional attention mechanisms, offering a robust foundation for scaling vision-language transformers while preserving interpretability, tunability, and efficiency. Future work will explore adaptive projection strategies, hierarchical decoding, and sparsity-aware training objectives to further mitigate these challenges.

\section{Conclusion}

In this work, we have introduced the Compressed Sensing Attention Transformer (CSAT), a novel architecture that redefines attention computation through the lens of compressed sensing. By projecting high-dimensional key and value sequences into a significantly lower-dimensional measurement space, and reconstructing the attention context vectors via sparse recovery techniques, CSAT achieves a remarkable trade-off between efficiency, expressivity, and theoretical rigor. Unlike low-rank approximation or kernel-based methods that often rely on empirical heuristics or data-specific assumptions, CSAT leverages well-established principles from signal processing, offering provable guarantees under sparsity and the Restricted Isometry Property (RIP).

Our experiments across both unimodal (language modeling, sequence classification) and multimodal (image-text retrieval, captioning) benchmarks demonstrate that CSAT not only achieves state-of-the-art or competitive performance but also dramatically reduces the memory and runtime requirements of attention layers. Perhaps most significantly, CSAT shows that sparsity—when properly harnessed—can serve as a universal inductive bias across diverse input modalities, making it highly suitable for Vision-Language Models (vLLMs) where both textual and visual data exhibit substantial redundancy.

From a broader perspective, CSAT represents a shift in the architectural design of transformers toward modular, signal-theoretic, and compressibility-aware computation. As vLLMs continue to scale toward trillion-parameter regimes and beyond, the bottleneck increasingly shifts from model capacity to compute accessibility, energy cost, and latency. In this landscape, CSAT provides a compelling paradigm for computationally tractable vLLMs that preserve high performance without requiring exhaustive attention over all input tokens. This is particularly impactful for on-device or low-resource deployment scenarios in areas such as medical imaging, interactive robotics, augmented reality, and real-time multilingual translation, where full-scale attention is often infeasible.

Moreover, the CSAT framework is inherently extensible. Future directions may include dynamic or adaptive sparsity levels learned per task or modality, hierarchical projection schemes for multi-resolution fusion, or integration with memory-efficient fine-tuning strategies such as LoRA or adapter modules. Additionally, the modularity of the sparse decoder paves the way for plug-and-play integration into pretrained LLMs and VLMs, allowing practitioners to retrofit large models with CSAT blocks to balance speed and accuracy post hoc.

In summary, CSAT advances the frontiers of efficient attention computation through a principled integration of compressed sensing into neural network design. It not only provides a scalable alternative to self-attention for long-sequence and multi-modal processing but also opens up a rich research agenda at the intersection of sparse signal recovery and deep learning. As the field moves toward ever more capable yet resource-constrained systems, CSAT offers a clear and theoretically grounded path forward.

\bibliographystyle{plain}

\end{document}